\documentclass[a4paper,fleqn]{cas-dc}

\usepackage[authoryear,longnamesfirst]{natbib}

\def\tsc#1{\csdef{#1}{\textsc{\lowercase{#1}}\xspace}}
\tsc{QTTB}
\tsc{TNQT}
\tsc{TBP}
\tsc{NTH}

\begin{document}
\let\WriteBookmarks\relax
\def\floatpagepagefraction{1}
\def\textpagefraction{.001}

\shorttitle{HyMLRaman}    

\shortauthors{Q. T. T. Binh et~al.}  

\title [mode = title]{Hybrid Machine Learning-Assisted Raman Spectroscopy with Generative Feature Augmentation for Pharmaceutical Identification}  



%

\author[1,2]{Quach Thi Thai Binh}[]
\credit{Writing original draft, Visualization, Data curation, Methodology, Investigation, Formal analysis}
\affiliation[1]{Faculty of Physics and Physics Engineering, University of Science, Ho Chi Minh City 700000, Viet Nam}
\affiliation[2]{Vietnam National University, Ho Chi Minh City 700000, Viet Nam}

\author[1,2]{Ton Nu Quynh Trang}[]
\credit{Data curation, Formal analysis}

\author[2,3,4]{Thang B. Phan}[]
\credit{Resources, Funding acquisition}
\affiliation[3]{Advanced Materials Technology Institute Vietnam National University Ho Chi Minh City (formerly Affiliated with Center for Innovative Materials and Architectures), Ho Chi Minh City 700000, Viet Nam}
\affiliation[4]{University of Health Sciences (UHS), Viet Nam National University Ho Chi Minh City, Ho Chi Minh City 700000, Viet Nam}

\author[1,2]{Vu Thi Hanh Thu}[]
\credit{Supervision, Writing - review, Visualization, Investigation}
\cormark[1]
\ead{vththu@hcmus.edu.vn}

\author[5]{Nguyen Tuan Hung}[orcid=0000-0003-4156-6230]
\credit{Supervision, Writing - review, Validation, Methodology, Investigation, Formal analysis. All authors have given their approval for the final version of the manuscript.}
\affiliation[5]{Frontier Research Institute for Interdisciplinary Sciences, Tohoku University, Sendai 980-8578, Japan}
\cormark[2]
\ead{nguyen.tuan.hung.e4@tohoku.ac.jp}
\cortext[cor1]{Corresponding author}
\cortext[cor2]{Principal corresponding author}

\begin{abstract}
Rapid and reliable identification of pharmaceutical residues is important for safeguarding public health, ensuring food safety, and enabling practical Raman-based screening. In this study, we propose HyMLRaman, a hybrid Raman spectroscopy framework that combines deep spectral feature extraction, generative models, and classical machine-learning classifiers to identify six pharmaceutical compounds, including amoxicillin, chloramphenicol, ciprofloxacin, tetracycline, ibuprofen, and paracetamol. Raman spectra are converted into spectral images and encoded with several deep neural-network backbones, among which EfficientNet-B3 yields the most effective representation. The resulting 1536-dimensional embeddings are then used to train downstream classifiers, including SVM, KNN, logistic regression, random forest, XGBoost, and ANN, using stratified 10-fold cross-validation. The hybrid EfficientNet-B3--SVM configuration achieves the strongest baseline performance, reaching 96.31\% accuracy and a macro-F1 score of 96.36\%, outperforming the standalone CNN baseline. To address limited-data conditions, a generative model, a DDPM-based feature augmentation, is introduced in a PCA-reduced EfficientNet-B3 latent space. The low-data ablation results show that DDPM augmentation provides selective benefits, particularly for KNN with reduced training fractions, and that its effect remains classifier-dependent. Finally, an application-level Raman Pharmaceutical Analyzer demonstrates the feasibility of embedding the trained model into an interactive Raman analysis workflow. These results suggest that HyMLRaman provides a practical and interpretable route for rapid Raman-based pharmaceutical screening.
\end{abstract}





\maketitle

\section{INTRODUCTION}
The growing presence of pharmaceutical residues in food, water, and clinical samples has raised increasing public-health concerns, motivating the need for rapid identification of widely used drugs such as amoxicillin (AMX), chloramphenicol (CHL), tetracycline (TET), ciprofloxacin (CIP), ibuprofen (IBU), and paracetamol (PAR)~\cite{Patel2019, Maghsodian2022, Gyesi2022}. Raman spectroscopy provides a fast, non-destructive analytical tool with molecule-specific vibrational signatures, making it an appealing solution for on-site screening~\cite{Liu2025, Chandra2024}. However, the spectral variability encountered in practical measurements often challenges traditional chemometric approaches and motivates the integration of modern machine learning techniques~\cite{Liu2025,Liu2026, Pocheville2025}.

In response to these challenges, machine learning (ML) techniques, including contemporary neural-network–based models, have been increasingly coupled with Raman spectroscopy~\cite{Liu2025, Liu2026, Wang2024, Sunil2024,Luo2022, Rashidi2025, Zarei2023,Wang2025}. Classical ML algorithms such as large-margin methods (e.g., support vector machines, SVMs) and random forests can handle complex decision boundaries but typically rely on carefully engineered features extracted from one-dimensional spectra~\cite{Liu2025, Liu2026, Pocheville2025, Sunil2024, Rahman2024}. In contrast, convolutional neural networks (CNNs) are specifically designed to learn hierarchical representations from spectral images, enabling the extraction of subtle structures that are not easily captured by manually constructed descriptors~\cite{Liu2025, Liu2026, Wang2024, Rieger2023, Binh2025}. Several recent investigations have reported that CNN-based feature encoders, when combined with powerful ensemble classifiers such as the gradient boosting algorithm (XGBoost), can substantially improve predictive accuracy and robustness compared with CNN-only or purely classical pipelines~\cite{Binh2025, Jiao2021, Zou2025}. Despite these advances, most existing Raman ML studies adopt a single modeling paradigm (either classical ML or deep learning), thereby underutilizing the complementary strengths of both~\cite{Liu2026, Luo2022, Binh2025, Jiao2021}. This motivates a hybrid strategy that combines the representational power of deep networks with the decision robustness and regularization benefits of classical classifiers~\cite{Rieger2023, Binh2025, Zou2025, Wjcik2026}.

\begin{figure*}[t!]
    \centering
    \includegraphics[width=0.95\linewidth]{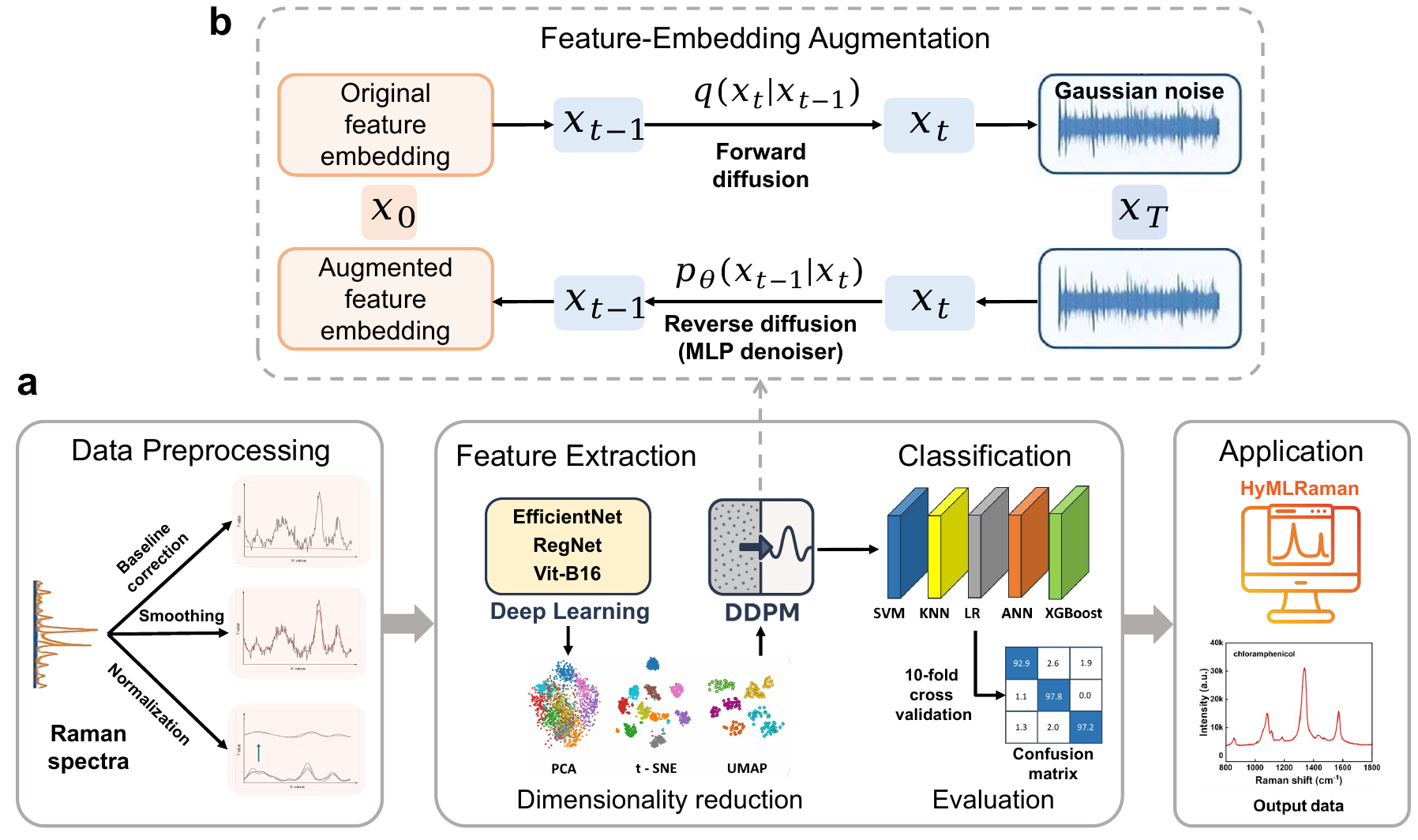}
    \caption{Schematic overview of the proposed HyMLRaman framework. (a) Cross-validation pipeline: (1) the data preprocessing, in which interpolated spectra are reshaped into $128 \times 128$ spectral images, (2) feature embeddings (1536-D) are extracted using several ML architectures (EfficientNet-B0, EfficientNet-B3, RegNetY-400M, and ViT-B16), and multiple classifiers (SVM, KNN, LR, RF, ANN, XGBoost) are evaluated under stratified 10-fold cross-validation with visualization and confusion-matrix analysis, (3) an end-to-end application is built based on the trained EfficientNet-B3 classifier. (b) DDPM-based diffusion augmentation for generating feature embeddings via forward noise injection and learned reverse denoising.}
    \label{fig:model}
\end{figure*}

In this work, we introduce HyMLRaman, a hybrid machine-learning framework for rapid identification of six pharmaceutical compounds from Raman spectra. A total of 1003 Raman spectral images are collected from publicly accessible datasets acquired under comparable measurement conditions. This dataset is used to fine-tune an EfficientNet-B3 encoder, whose embeddings serve as a unified feature space for downstream classifiers, including SVM, $k$-nearest neighbors (KNN), logistic regression, random forest, XGBoost, and a shallow neural network. These hybrid pipelines achieve accuracies ranging from 94.62\% to 96.31\% across the evaluated downstream classifiers, with the EfficientNet-B3--SVM configuration providing the strongest baseline performance of 96.31\% accuracy and 96.36\% macro-F1. To further stabilize learning with limited data, a denoising diffusion probabilistic model (DDPM) is trained in the embedding space to perform feature-level augmentation. Furthermore, by comparing the CNN and Vision Transformer (ViT-B16) models, we also highlight the localized peak characteristics in Raman spectra. Finally, an end-to-end EfficientNet-B3 classifier integrated into a PyQt interface demonstrates the practical feasibility of real-time Raman-based drug screening.

\begin{figure*}[t]
    \centering
    \includegraphics[width=0.85\linewidth]{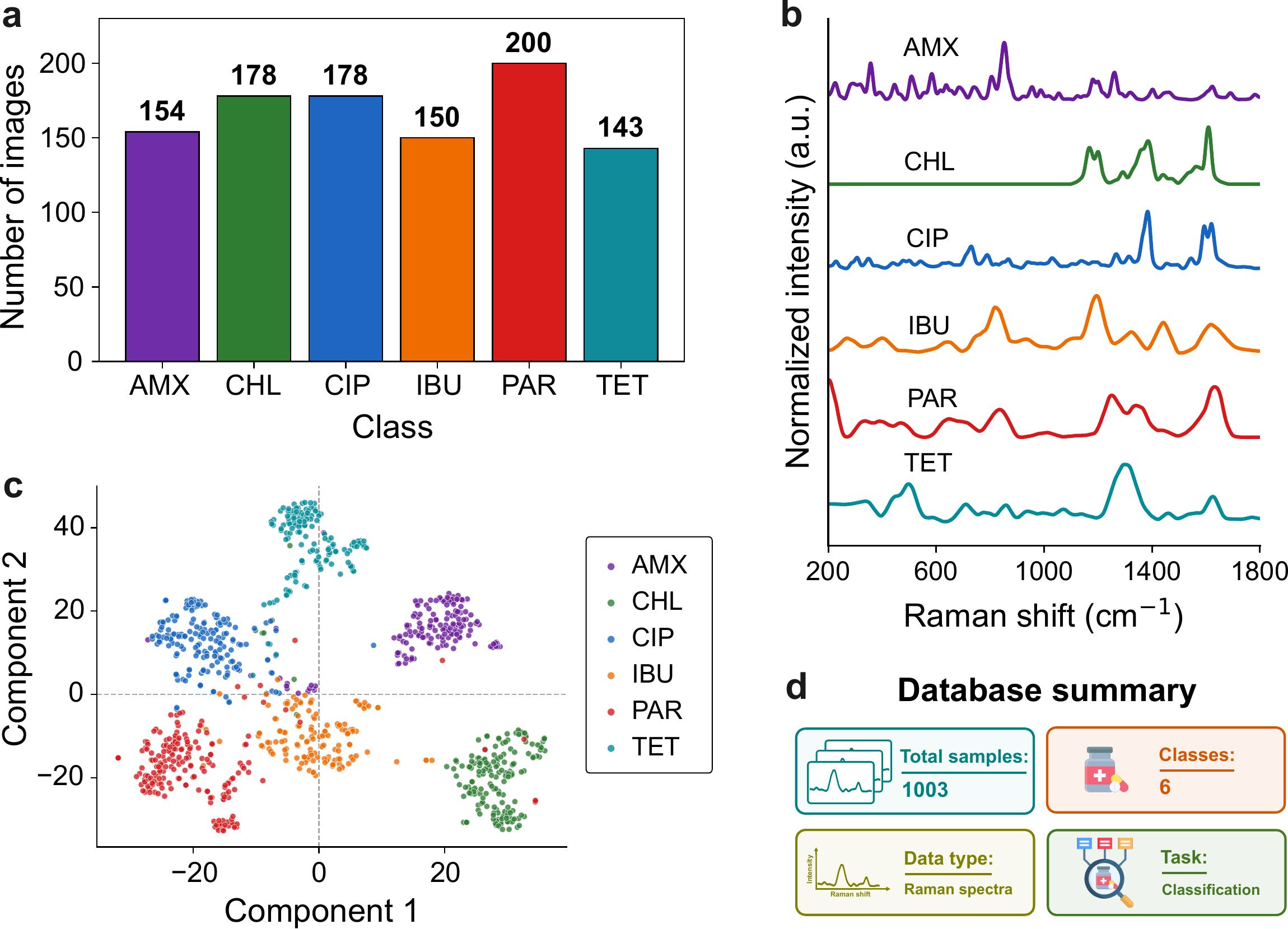}
    \caption{Overview of the Raman spectral dataset.
    (a) Number of Raman spectral images in each pharmaceutical class: amoxicillin (AMX), chloramphenicol (CHL), ciprofloxacin (CIP), ibuprofen (IBU), paracetamol (PAR), and tetracycline (TET).
    (b) Representative normalized Raman spectra of the six pharmaceutical compounds, vertically offset for clarity.
    (c) t-SNE visualization of the EfficientNet-B3 feature embeddings for the six pharmaceutical classes.
    (d) Summary of the dataset, including the total number of samples, number of classes, data type, and classification task.}
    \label{fig:data}
\end{figure*}

\section{METHODS AND MATERIALS}
\subsection{HyMLRaman: A Hybrid Machine Learning Framework for Raman Spectra}
In Figure~\ref{fig:model}a, we illustrate the HyMLRaman framework used in this study. The HyMLRaman framework begins with the Raman spectral dataset and preprocessing, then proceeds to feature extraction with several deep neural-networks models, including the CNN models with EfficientNet-B0, EfficientNet-B3, and RegNetY-400M, and the vision transformer model with ViT-B16, followed by dimensionality reduction and classification using several ML models, including SVM, KNN, LR, RF, ANN, and XGBoost. The performance of the HyMLRaman framework is measured with accuracy, F1-score, and a confusion matrix. The framework ends with a user interface, developed to enable practical use of the model for the rapid detection of pharmaceutical residues from Raman spectra. To improve the limited dataset, a generative model based on the DDPM is adopted for feature augmentation from the dimensionality-reduced space, as shown in Figure~\ref{fig:model}b. All code and the user interface application are available for download on the GitHub page: \url{https://github.com/nguyen-group/HyMLRaman}.

\subsection{Raman Spectral Acquisition and Image Construction}
Raman spectra of six pharmaceutical compounds are collected from publicly accessible datasets acquired under comparable measurement conditions, yielding a total of 1003 samples, as shown in Figure~\ref{fig:data}a. In the implemented pipeline, the spectral data are prepared as Raman spectral images and resized to the deep neural-networks (CNN and transformer models) input resolution by applying additional Raman-specific baseline correction, despiking, or smoothing. For deep neural network architectures, the images are converted to a three-channel grayscale format and normalized using ImageNet statistics, enabling the pretrained deep neural network backbones to extract localized Raman peak patterns. The dataset is summarized in Figure~\ref{fig:data}d and is available on the GitHub page: \url{https://github.com/nguyen-group/HyMLRaman/data}.

\subsection{Feature Extraction from Deep Neural-Networks}
A comprehensive evaluation of several state-of-the-art ML architectures is conducted to identify an optimal feature extractor for Raman spectral images, including the CNN models (EfficientNet-B0, EfficientNet-B3, and RegNetY-400M) and the transformer model (Vision Transformer ViT-B16). Here, RegNetY-400M and EfficientNet-B0 serve as architectural baselines that probe different depth--width regimes, ensuring that the observed performance trends are attributable to structural advantages rather than mere increases in parameter count. This comparative design provided a rigorous basis for selecting an optimal backbone for Raman-specific embeddings.

To complement the CNN-based backbones, ViT-B16 is included as a representative
patch-based self-attention architecture. Compared with convolutional networks,
Vision Transformers possess weaker built-in spatial locality biases and model
interactions between image patches primarily through self-attention mechanisms.
For Raman spectral classification, however, chemically informative information is
often concentrated in localized vibrational bands and peak structures. Previous
Raman studies have shown that both convolutional and transformer-based models can
learn discriminative spectral representations, although their relative performance
depends on the spectral representation and available training data~\cite{Wang2024,Rieger2023}.
In the present dataset, the convolutional backbones outperform ViT-B16, suggesting
that the locality-preserving characteristics of CNNs are advantageous for the
Raman spectral-image representation used in this work.

Across the evaluated architectures, EfficientNet-B3 achieves the strongest classification performance and is therefore selected as the feature extractor for the subsequent analyses. Accordingly, the 1536-dimensional activations obtained from its global average pooling layer are adopted as the unified feature representation for all downstream classifiers.
As shown in Figure~\ref{fig:data}c, the EfficientNet-B3 embeddings are visualized using t-distributed stochastic neighbor embedding (t-SNE)~\cite{vandermaaten2008tsne}.
The resulting two-dimensional projection exhibits six visually distinguishable class regions, suggesting that the learned EfficientNet-B3 representation contains substantial class-discriminative information for the six pharmaceutical compounds.

\subsection{Machine Learning Classifiers}
The 1536-dimensional EfficientNet-B3 embeddings are used as inputs to multiple supervised classifiers. Let $\mathbf{x}_i \in \mathbb{R}^{1536}$ denote the embedding extracted from the $i$-th Raman spectral image, and let $y_i \in \{1,\ldots,M\}$ denote the corresponding class label, where $M$ is the number of pharmaceutical classes and $i=1,\ldots,N$. When required by a downstream classifier, feature standardization is fitted only on the training split within each cross-validation fold and then applied to the corresponding validation split. This fold-wise procedure prevents information leakage during model evaluation.

For SVM classification, a radial basis function (RBF) kernel is employed following the standard support-vector formulation~\cite{Cortes1995}:
\begin{equation}
f(\mathbf{x})=\sum_{i=1}^{N}\alpha_i y_i K(\mathbf{x}_i,\mathbf{x})+b,
\end{equation}
where the RBF kernel is defined as
$K(\mathbf{x}_i,\mathbf{x})=\exp\!\left(-\gamma\|\mathbf{x}_i-\mathbf{x}\|^2\right)$.
Here, $\mathbf{x}$ is the query embedding to be classified, $\mathbf{x}_i$ is the
$i$-th training embedding, $\alpha_i$ is the learned coefficient associated with the $i$-th support vector, $y_i$ is the class label of the $i$-th training sample, and $b$ is the bias term. The parameter $\gamma$ controls the width of the RBF kernel and thus the locality of the decision boundary. In the present implementation, $\gamma$ is optimized by grid search using the training data.

For the KNN classifier, the predicted label of a query embedding is determined by
majority voting among its $k$-nearest neighbors~\cite{cover1967nearest}:
\begin{equation}
\hat{y}=
\operatorname{mode}\!\left(
\{y_i\}_{i\in\mathcal{N}_k(\mathbf{x})}
\right),
\end{equation}
where $\hat{y}$ is the predicted class label, $\mathcal{N}_k(\mathbf{x})$ denotes the index set of the $k$-nearest training samples to $\mathbf{x}$ under the selected distance metric, and $\{y_i\}_{i\in\mathcal{N}_k(\mathbf{x})}$ is the set of class labels associated with those neighbors. The neighborhood size $k$ is selected by cross-validation using the training data.

For multinomial logistic regression~\cite{bohning1992multinomial}, the posterior
probability that a sample belongs to class $k$ is estimated using the softmax function:
\begin{equation}
P(y=k\mid\mathbf{x})=
\frac{\exp\!\left(\mathbf{w}_k^{T}\mathbf{x}+b_k\right)}
{\sum_{j=1}^{K}\exp\!\left(\mathbf{w}_j^{T}\mathbf{x}+b_j\right)},
\end{equation}
where $P(y=k\mid\mathbf{x})$ is the predicted probability that embedding $\mathbf{x}$ belongs to class $k$, $\mathbf{w}_k$ is the weight vector associated with class $k$, and $b_k$ is the corresponding bias term. The denominator sums the unnormalized scores over all $K$ classes to produce a valid probability distribution. Model parameters are estimated with $L_2$ regularization, and the regularization strength is tuned using the training data.

Random forest (RF) is additionally evaluated as an ensemble-based classifier operating on the EfficientNet-B3 embeddings. Multiple decision trees are constructed from resampled training data and subsets of candidate features, and the final class prediction is obtained by aggregating the predictions of the individual trees. RF is included to provide a tree-ensemble baseline with nonlinear decision capability. XGBoost is also evaluated as a gradient-boosted tree classifier. In contrast to RF, where multiple trees contribute to an aggregated ensemble prediction, XGBoost constructs trees sequentially, with successive learners optimized to improve the objective of the preceding ensemble. Tree-based ensemble approaches, including RF and XGBoost, have demonstrated strong performance in vibrational-spectroscopy classification and feature-learning applications~\cite{daSilva2025Tree,Zou2025}.

A multilayer perceptron (MLP) is also trained directly on the embedding vectors. The MLP includes two fully connected hidden layers with 256 and 128 neurons, respectively, each followed by rectified linear unit (ReLU) activation and dropout ($p=0.3$) ~\cite{srivastava2014dropout}. The output layer uses softmax activation to generate class probabilities over the $K$ target classes, and the network parameters are optimized using Adam.

In addition to the classical machine-learning classifiers, an embedding-level one-dimensional CNN is implemented as a neural baseline operating on the same EfficientNet-B3 feature representation. In this case, the input is not the raw Raman spectrum but the 1536-dimensional embedding vector extracted from the Raman spectral image. Let $\mathbf{x}_i \in \mathbb{R}^{1536}$ denote the input embedding. The embedding is treated as a one-dimensional feature sequence and passed through convolutional blocks to capture local dependencies among neighboring embedding components:
\begin{equation}
\mathbf{F}^{(\ell)} =
\sigma\left(
\mathrm{Conv1D}
\left(
\mathbf{F}^{(\ell-1)}; \mathbf{W}^{(\ell)}, \mathbf{b}^{(\ell)}
\right)
\right),
\quad
\mathbf{F}^{(0)} = \mathbf{x}_i ,
\end{equation}
where $\mathbf{F}^{(\ell-1)}$ and $\mathbf{F}^{(\ell)}$ denote the input and output feature maps of the $\ell$-th convolutional block, respectively, and $\mathbf{W}^{(\ell)}$ and $\mathbf{b}^{(\ell)}$ are the learnable convolution kernel and bias. The activation function is denoted by $\sigma(\cdot)$. This embedding-level CNN provides a neural-network baseline within the same 1536-dimensional EfficientNet-B3 feature space used by the downstream classifiers.

After the final convolutional block, the extracted representation is mapped to class probabilities as
\begin{equation}
\hat{\mathbf{p}} =
\mathrm{softmax}
\left(
\mathbf{W}_c\mathbf{z}+\mathbf{b}_c
\right),
\end{equation}
where $\mathbf{z}$ is the compact latent representation produced by the convolutional network, $\hat{\mathbf{p}}\in\mathbb{R}^{K}$ is the predicted class-probability vector, and $K$ is the number of target classes. The final predicted label is determined as $\hat{y}=\arg\max_k \hat{p}_k$.

\subsection{DDPM-Based Feature-Embedding Augmentation}
We use a DDPM-based module for feature-level augmentation in the limited-data experiments. The module operates on EfficientNet-B3 embeddings and does not generate raw Raman spectra. For each Raman spectral image, the fine-tuned EfficientNet-B3 encoder produces a 1536-dimensional embedding $\mathbf{x}_i \in \mathbb{R}^{1536}$ with class label $y_i \in \{1,\ldots,K\}$. DDPM generation is then performed in a PCA-reduced latent version of this feature space. This design follows recent diffusion-based augmentation strategies for Raman and spectroscopic learning under limited-data conditions~\cite{Yao2025DiffRaman}, while restricting generation to the learned feature domain rather than the original Raman intensity domain. Within each cross-validation fold, all preprocessing and DDPM operations are fitted only on the training split. The validation fold is not used for standardization, PCA fitting, DDPM training, candidate generation, or filtering.

The training embeddings are standardized and projected into a PCA latent space ~\cite{Jolliffe2016PCA}:
\begin{equation}
    \mathbf{z}_i =
    \mathbf{P}^{\top}
    \left(
    \frac{\mathbf{x}_i-\boldsymbol{\mu}_{\mathrm{tr}}}
    {\boldsymbol{\sigma}_{\mathrm{tr}}}
    \right),
    \qquad
    \mathbf{z}_i \in \mathbb{R}^{d},
\end{equation}
where $\boldsymbol{\mu}_{\mathrm{tr}}$ and $\boldsymbol{\sigma}_{\mathrm{tr}}$ are computed from the training embeddings, $\mathbf{P}$ is the PCA projection matrix fitted within the same fold, and $d=64$. This PCA-reduced feature space is used as the input domain for DDPM training and generation.

The DDPM is trained on the original latent features $\mathbf{z}_0$.
Following the standard discrete-time DDPM formulation~\cite{Chang2026Diffusion},
the forward diffusion process progressively adds Gaussian noise according to
\begin{equation}
    q(\mathbf{z}_t|\mathbf{z}_{t-1}) =
    \mathcal{N}
    \left(
    \sqrt{1-\beta_t}\mathbf{z}_{t-1},
    \beta_t\mathbf{I}
    \right),
\end{equation}
where $\beta_t$ denotes the variance schedule. Defining
$\alpha_t=1-\beta_t$ and
$\bar{\alpha}_t=\prod_{s=1}^{t}\alpha_s$,
the latent feature at timestep $t$ can be sampled directly from
$\mathbf{z}_0$ as
\begin{equation}
    \mathbf{z}_t =
    \sqrt{\bar{\alpha}_t}\mathbf{z}_0
    +
    \sqrt{1-\bar{\alpha}_t}\boldsymbol{\epsilon},
    \qquad
    \boldsymbol{\epsilon}\sim\mathcal{N}(\mathbf{0},\mathbf{I}).
\end{equation}
A class-conditional denoising network
$\boldsymbol{\epsilon}_{\theta}(\mathbf{z}_t,t,y)$ is trained to predict the
injected noise from the noisy latent feature $\mathbf{z}_t$, the timestep $t$,
and the class label $y$, following the conditional latent-diffusion paradigm
used for Raman data generation~\cite{Yao2025DiffRaman}.
In our implementation, timestep and class embeddings are concatenated with
the noisy latent feature and provided to the denoising network. The model is
optimized using the standard DDPM noise-prediction objective~\cite{Chang2026Diffusion}:
\begin{equation}
    \mathcal{L}_{\mathrm{DDPM}} =
    \mathbb{E}_{\mathbf{z}_0,y,t,\boldsymbol{\epsilon}}
    \left[
    \left\|
    \boldsymbol{\epsilon}
    -
    \boldsymbol{\epsilon}_{\theta}(\mathbf{z}_t,t,y)
    \right\|_2^2
    \right].
\end{equation}

After training, the DDPM generates class-conditioned latent candidates from Gaussian noise. These candidates are not directly added to the training set. We first screen them by classifier confidence and nearest-neighbor consistency. A balanced logistic-regression classifier is fitted on the original latent features of the training split. A generated candidate $\tilde{\mathbf{z}}$ conditioned on class $c$ is retained only if
\begin{equation}
    \hat{y}(\tilde{\mathbf{z}})=c,
    \qquad
    P(y=c|\tilde{\mathbf{z}})\geq \tau ,
\end{equation}
where $\tau=0.85$. The candidate must also satisfy a nearest-neighbor distance constraint:
\begin{equation}
    d(\tilde{\mathbf{z}},\mathcal{Z}_c)
    =
    \min_{\mathbf{z}_j \in \mathcal{Z}_c}
    \|\tilde{\mathbf{z}}-\mathbf{z}_j\|_2
    \leq d_c^{\mathrm{thr}},
\end{equation}
where $\mathcal{Z}_c$ is the set of original latent features from class $c$. The threshold is computed from the original within-class nearest-neighbor distances:
\begin{equation}
    d_c^{\mathrm{thr}} =
    \lambda \cdot
    \mathrm{Percentile}_{95}
    \left(
    \left\{
    \min_{\mathbf{z}_j \in \mathcal{Z}_c,\,j\neq i}
    \|\mathbf{z}_i-\mathbf{z}_j\|_2
    \right\}
    \right),
\end{equation}
with $\lambda=1.25$.

The accepted DDPM-generated features form $\mathcal{D}_{\mathrm{DDPM}}$ and are concatenated with the original training features:
\begin{equation}
    \mathcal{D}_{\mathrm{aug}} =
    \mathcal{D}_{\mathrm{original}}
    \cup
    \mathcal{D}_{\mathrm{DDPM}} .
\end{equation}
Downstream classifiers are trained on either $\mathcal{D}_{\mathrm{original}}$ or $\mathcal{D}_{\mathrm{aug}}$, and evaluation is performed only on the untouched validation fold. The effect of DDPM augmentation is evaluated in the low-data ablation study using 25\%, 50\%, 75\%, and 100\% of the training data, with the average numbers of original and accepted DDPM-generated samples listed in Table S3.

\subsection{Cross-Validation Strategy}

A stratified ten-fold cross-validation protocol is used to preserve comparable class distributions across folds and to obtain fold-wise estimates of predictive performance under repeated train--validation partitioning~\cite{Varoquaux2017CV}.
In each fold, the classifier is trained on nine folds and evaluated on the remaining fold, and the reported results are averaged over the ten fold-wise estimates. Accuracy, macro-precision, macro-recall, and macro-F1 score are used as the main evaluation metrics for multiclass Raman identification. Confusion matrices and one-vs-rest ROC--AUC curves are additionally reported for selected classifiers. All cross-validation experiments use a fixed random seed (SEED = 1337) to ensure consistent data partitioning.

\subsection{Implementation Details}
All spectral processing, deep feature extraction, diffusion modeling, and classifier training routines are implemented in Python. Deep learning models, including EfficientNet, RegNet, and ViT, and the DDPM architecture, are implemented in PyTorch, whereas classical machine-learning models are implemented in scikit-learn. Training is performed on an NVIDIA GPU workstation. The computational efficiencies are listed in Tables S1 and S2 for NVIDIA GeForce RTX 3060 and 4090. Hyperparameters such as learning rate, batch size, number of training epochs, and augmentation ratios are selected via preliminary validation experiments. All random seeds, model configurations, and preprocessing scripts are archived to enable complete reproducibility of the experimental pipeline.

\begin{figure*}[t!]
    \centering
    \includegraphics[width=0.85\linewidth]{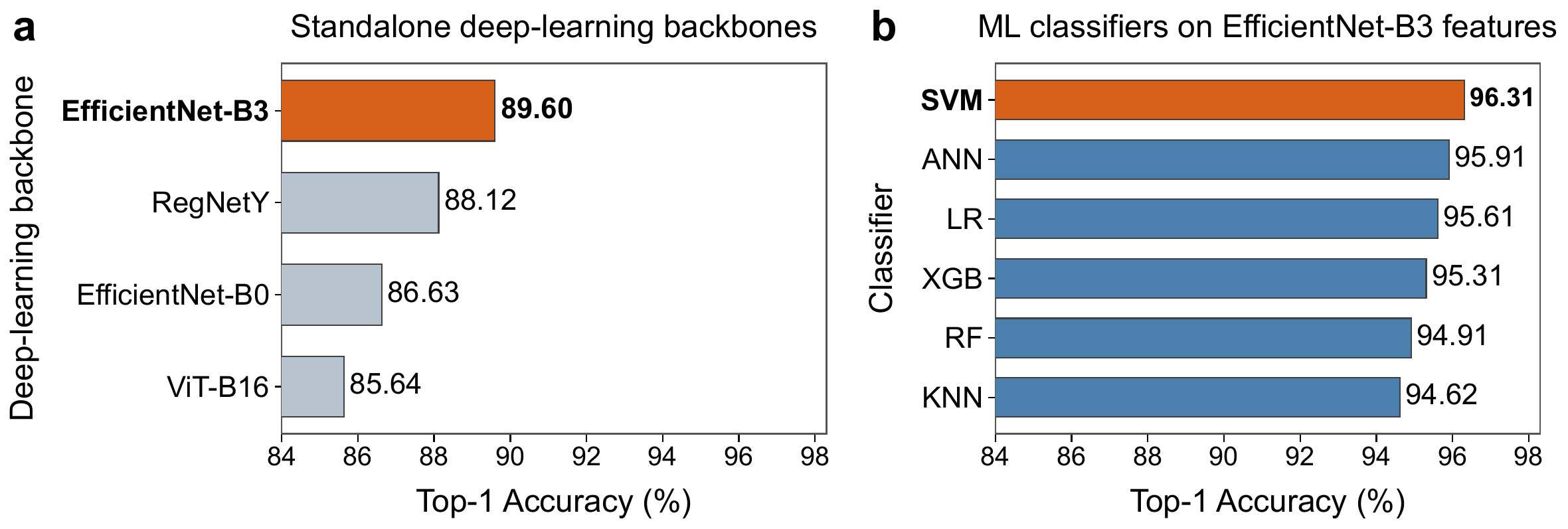}
    \caption{
    Top-1 accuracy comparison between standalone deep learning backbones and classical machine-learning classifiers trained on EfficientNet-B3 embeddings. 
    The orange bar highlights the best-performing model, SVM, which achieves 96.31\% accuracy. 
    Deep learning backbones are shown in light gray, while classical machine-learning classifiers are shown in blue.
    }
    \label{fig:overall_performance}
\end{figure*}

\subsection{Experimental Raman Measurement}

Experimental Raman/SERS measurements are conducted using a Horiba XploRA PLUS
Raman system equipped with a 50$\times$ objective lens and a 532~nm excitation
laser, following the measurement protocol reported by Trang et al.~\cite{trang2023hotspot}.
Four pharmaceutical compounds, namely paracetamol, tetracycline, ciprofloxacin,
and ibuprofen, are experimentally measured as external test samples for evaluating
the practical identification capability of the ML-assisted Raman framework.

All samples are dried under ambient conditions at room temperature prior to
measurement. The Raman/SERS signals are subsequently collected from the dried
sample spots, with the laser focused to an approximately 2~$\mu$m spot diameter.
These experimentally measured spectra are independent of the data used for model
training and are employed only for the application-level demonstration of the
proposed framework.

\section{RESULTS AND DISCUSSION}  

\subsection{Cross-Validation Consistency}

The stratified ten-fold split produces nearly identical validation proportions across all folds. Because the dataset contains 1003 Raman samples, the validation folds contain either 100 or 101 samples, corresponding to a narrow range of 9.97--10.07\%. The average validation proportion is 10.00\%, with a standard deviation of only 0.048\%, as shown in Figure S1. This confirms that the evaluation is not influenced by large fold-size variation. More importantly, the stratified design maintains comparable class representation across the ten validation folds. Therefore, the reported cross-validation metrics reflect classifier behavior across the full Raman dataset rather than the outcome of a single favorable train--validation split. All subsequent performance values are reported as averages over the ten validation folds. 

\subsection{Overall Performance of CNN--ML Classifiers}

Figures~\ref{fig:overall_performance}a and~\ref{fig:overall_performance}b show the Top-1 accuracy of standalone deep learning backbones and classical machine-learning classifiers trained on EfficientNet-B3 embeddings, respectively. Among the standalone CNN/transformer backbones, EfficientNet-B3 gives the best result, with an accuracy of 89.60\%, while RegNetY, EfficientNet-B0, and ViT-B16 achieve lower accuracies of 88.12\%, 86.63\%, and 85.64\%, respectively. Based on this comparison, EfficientNet-B3 is selected as the feature extractor for the subsequent hybrid CNN--ML classification pipeline. In Figure S2, we show the accuracy and loss of the EfficientNet-B3 backbone as a function of the number of epochs. The validation accuracy increases rapidly during the early epochs and stabilizes near 89\%, while the validation loss remains low after convergence. Next, we use the EfficientNet-B3 embeddings as input features for classical classifiers. The accuracy increases to above 94\% for all six downstream classifiers, as shown in Figure~\ref{fig:overall_performance}b. This indicates that the extracted embeddings form a suitable feature representation for downstream pharmaceutical identification. SVM achieves the highest accuracy of 96.31\%, followed by ANN, LR, and XGBoost with accuracies of 95.91\%, 95.61\%, and 95.31\%, respectively. RF and KNN remain slightly lower but still comparable, reaching 94.91\% and 94.62\%.

Overall, the hybrid CNN--ML configuration performs better than the standalone deep learning backbones in this dataset. These results suggest that EfficientNet-B3 is more effective as a feature extractor than as a standalone classifier. The best performance obtained by SVM further indicates that the EfficientNet-B3 feature space is well suited to margin-based classification for multi-class pharmaceutical identification.

\subsection{Baseline Performance and Low-Data DDPM Ablation}

\begin{table}[t]
\centering
\definecolor{tableblue}{HTML}{0B4F8A}
\definecolor{headergray}{HTML}{E6E6E6}
\noindent\textcolor{tableblue}{\rule{\linewidth}{1.1pt}}
\caption{Baseline performance of the hybrid EfficientNet-B3--ML classifiers without DDPM augmentation, in which ML includes SVM, ANN, LR, XGB, RF, KNN, and CNN models.}
\label{tab:org_baseline}
\vspace{3pt}
\renewcommand{\arraystretch}{1.15}
\setlength{\tabcolsep}{5pt}
\small
\resizebox{\linewidth}{!}{
\begin{tabular}{lcccccc}
\rowcolor{headergray}
\textbf{Hybrid ML model} &
\textbf{Acc. (\%)} &
\textbf{Prec. (\%)} &
\textbf{Rec. (\%)} &
\textbf{F1 (\%)} &
\textbf{Acc. std.} &
\textbf{F1 std.} \\
EfficientNet-B3--SVM      & \textbf{96.31} & \textbf{96.57} & \textbf{96.35} & \textbf{96.36} & 2.14 & 2.12 \\
EfficientNet-B3--ANN      & 95.91 & 96.19 & 95.91 & 95.93 & 1.61 & 1.61 \\
EfficientNet-B3--LR       & 95.61 & 95.86 & 95.58 & 95.62 & 2.14 & 2.15 \\
EfficientNet-B3--XGB      & 95.31 & 95.69 & 95.43 & 95.42 & 1.85 & 1.86 \\
EfficientNet-B3--RF       & 94.91 & 95.26 & 94.99 & 94.97 & 1.60 & 1.55 \\
EfficientNet-B3--KNN      & 94.62 & 95.03 & 94.64 & 94.66 & 1.75 & 1.77 \\
EfficientNet-B3--CNN      & 92.62 & 92.95 & 92.45 & 92.53 & 2.46 & 2.59 \\
\end{tabular}
}
\noindent\textcolor{tableblue}{\rule{\linewidth}{1.1pt}}
\end{table}

The hybrid EfficientNet-B3--ML results provide the reference performance before introducing the DDPM augmentation branch. As shown in Table~\ref{tab:org_baseline}, SVM achieves the strongest baseline performance on the original 1536-dimensional EfficientNet-B3 embeddings, with an accuracy of 96.31\% and a macro-F1 score of 96.36\%. ANN and LR also perform strongly, reaching accuracies of 95.91\% and 95.61\%, respectively, followed by XGBoost at 95.31\%. RF and KNN remain slightly lower but still comparable, with accuracies of 94.91\% and 94.62\%. In comparison, the embedding-level CNN baseline trained on the same EfficientNet-B3 feature representation reaches 92.62\% accuracy and 92.53\% macro-F1 score, which is lower than all classical machine-learning classifiers trained on EfficientNet-B3 embeddings. This result supports using EfficientNet-B3 as a feature extractor, followed by a separate downstream classifier for final decision-making.

For the strongest classifiers, accuracy, macro-precision, macro-recall, and macro-F1 remain close to each other. This indicates that the reported performance is not driven by a single dominant class, but is maintained across the six pharmaceutical compounds. Among the tested classifiers, the hybrid EfficientNet-B3--SVM provides the strongest baseline and therefore serves as the main reference for the overall classification framework.

\begin{table}[tp]
\centering
\definecolor{tableblue}{HTML}{0B4F8A}
\definecolor{headergray}{HTML}{E6E6E6}
\noindent\textcolor{tableblue}{\rule{\linewidth}{1.1pt}}
\caption{Low-data ablation of DDPM-based feature augmentation using PCA-reduced EfficientNet-B3 latent features. Accuracy and macro-F1 values
are reported as mean percentages over stratified ten-fold cross-validation. Positive $\Delta$ values indicate improvement after DDPM-based feature augmentation.}
\label{tab:ddpm_ablation}
\vspace{3pt}
\renewcommand{\arraystretch}{1.15}
\setlength{\tabcolsep}{3.2pt}
\small
\resizebox{\linewidth}{!}{
\begin{tabular}{clcccccc}
\rowcolor{headergray}
\textbf{Frac.} &
\textbf{Model} &
\textbf{Original Acc.} &
\textbf{DDPM Acc.} &
$\boldsymbol{\Delta}$\textbf{ Acc.} &
\textbf{Original F1} &
\textbf{DDPM F1} &
$\boldsymbol{\Delta}$\textbf{ F1} \vspace{3pt} \\
\hline
25\%  & KNN & 93.616 & 94.115 & \textbf{+0.499} & 93.628 & 94.154 & \textbf{+0.526} \\
25\%  & SVM & 94.510 & 94.811 & +0.301 & 94.582 & 94.861 & +0.279 \\
25\%  & RF  & 93.715 & 93.714 & -0.001 & 93.737 & 93.756 & +0.019 \\
25\%  & LR  & 94.311 & 94.212 & -0.099 & 94.356 & 94.235 & -0.121 \\
25\%  & XGB & 93.119 & 92.919 & -0.200 & 93.179 & 92.902 & -0.277 \\
25\%  & ANN & 94.810 & 94.210 & -0.600 & 94.838 & 94.225 & -0.613 \\
\hline
50\%  & KNN & 93.918 & 94.713 & \textbf{+0.795} & 93.959 & 94.807 & \textbf{+0.848} \\
50\%  & SVM & 95.511 & 95.410 & -0.101 & 95.592 & 95.454 & -0.138 \\
50\%  & RF  & 93.915 & 93.815 & -0.100 & 93.987 & 93.809 & -0.178 \\
50\%  & LR  & 94.012 & 93.813 & -0.199 & 94.074 & 93.865 & -0.209 \\
50\%  & ANN & 94.813 & 94.514 & -0.299 & 94.859 & 94.604 & -0.256 \\
50\%  & XGB & 93.214 & 92.918 & -0.296 & 93.307 & 92.882 & -0.425 \\
\hline
75\%  & LR  & 94.015 & 94.315 & \textbf{+0.300} & 94.028 & 94.350 & \textbf{+0.322} \\
75\%  & KNN & 94.317 & 94.514 & +0.197 & 94.422 & 94.569 & +0.146 \\
75\%  & SVM & 95.514 & 95.612 & +0.098 & 95.579 & 95.653 & +0.074 \\
75\%  & ANN & 95.211 & 95.014 & -0.197 & 95.246 & 95.057 & -0.188 \\
75\%  & XGB & 94.114 & 93.415 & -0.699 & 94.200 & 93.409 & -0.791 \\
75\%  & RF  & 94.910 & 94.113 & -0.797 & 95.019 & 94.172 & -0.847 \\
\hline
100\% & SVM & 95.912 & 96.012 & \textbf{+0.100} & 95.949 & 96.035 & \textbf{+0.086} \\
100\% & RF  & 94.812 & 94.812 & +0.000 & 94.868 & 94.880 & +0.012 \\
100\% & LR  & 94.518 & 94.316 & -0.202 & 94.470 & 94.263 & -0.208 \\
100\% & KNN & 95.214 & 95.014 & -0.200 & 95.248 & 95.017 & -0.231 \\
100\% & XGB & 94.412 & 94.013 & -0.399 & 94.460 & 94.047 & -0.413 \\
100\% & ANN & 95.813 & 94.715 & -1.098 & 95.813 & 94.655 & -1.158 \\
\end{tabular}
}
\noindent\textcolor{tableblue}{\rule{\linewidth}{1.1pt}}
\end{table}

In Table~\ref{tab:ddpm_ablation}, we summarize the low-data ablation results of the DDPM branch. In this case, the 1536-dimensional EfficientNet-B3 embeddings are first standardized and reduced to a 64-dimensional PCA latent space within each fold. Classifiers trained only with original PCA-reduced features are then compared with classifiers trained using original latent features together with filtered DDPM-generated latent samples. The results show that the effect of DDPM augmentation is selective rather than uniform, depending on both the training-data fraction and the downstream classifier. 

At the 25\% training fraction, KNN and SVM show the clearest positive changes. KNN accuracy increases from 93.616\% to 94.115\%, while its macro-F1 score increases from 93.628\% to 94.154\%. SVM also improves from 94.510\% to 94.811\% accuracy and from 94.582\% to 94.861\% macro-F1. In contrast, LR shows a slight decrease after augmentation at the same fraction. At the 50\% training fraction, KNN again shows the largest improvement, with accuracy increasing by 0.795 percentage points and macro-F1 increasing by 0.848 percentage points. This suggests that the accepted DDPM-generated latent samples are particularly helpful for KNN when the original feature distribution is still sparsely sampled.

At higher training fractions, the effect becomes less pronounced. In the 75\% setting, LR, KNN, and SVM show small positive gains, with accuracy improvements of 0.300, 0.197, and 0.098 percentage points, respectively. With the full training set, SVM improves only slightly by 0.100 percentage points, whereas LR and KNN decrease slightly after augmentation. This trend indicates that DDPM-based augmentation is more useful when original training samples are limited, while its added value becomes weaker once the original latent distribution is better represented.

Overall, the original baseline and DDPM ablation results support two main findings. First, SVM is the strongest CNN--ML baseline on the original EfficientNet-B3 embeddings. Second, DDPM-generated features provide selective low-data benefits in the PCA-reduced latent space rather than a universal improvement across classifiers. The most consistent positive effect appears for KNN under reduced training-data fractions, with smaller gains for SVM in selected settings. Therefore, DDPM is best interpreted as an auxiliary feature-augmentation module for sparse-data conditions, not as a replacement for the original CNN--ML baseline.

\begin{figure}
    \centering
    \includegraphics[width=0.85\linewidth]{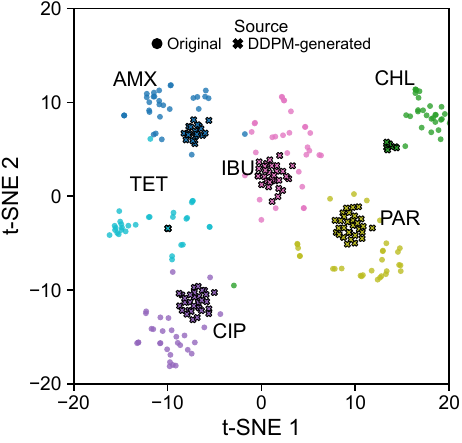}
    \caption{t-SNE visualization of original and filtered DDPM-generated latent features under the 25\% training-data setting. The original samples are shown as circles, whereas DDPM-generated samples are shown as cross markers. The synthetic samples are generated in the 64-dimensional PCA-reduced EfficientNet-B3 feature space and retained after classifier-confidence and nearest-neighbor filtering.}
    \label{fig:ddpm_tsne}
\end{figure}

\subsection{Feature-Space Consistency of DDPM-Generated Samples}

In Figure~\ref{fig:ddpm_tsne}, we show the t-SNE projection of original and DDPM-generated latent features under the 25\% training-data setting. For this visualization, we standardize the 1536-dimensional EfficientNet-B3 embeddings and reduce them to a 64-dimensional PCA latent space before DDPM training. 
The plotted data include 225 original training samples and 154 filtered DDPM-generated samples. As shown in Figure~\ref{fig:ddpm_tsne}, most accepted DDPM-generated samples lie within or close to the class regions formed by the original samples. This pattern is consistent with the confidence-based and nearest-neighbor filtering strategy, which removes generated candidates that deviate strongly from the original training distribution. Therefore, the DDPM samples are interpreted as additional local latent candidates rather than synthetic Raman spectra.

This visualization serves as a qualitative check of feature-space consistency. The proximity between original and DDPM-generated samples in the t-SNE map does not, by itself, demonstrate a performance gain. The quantitative effect of DDPM augmentation is evaluated separately through the low-data ablation results in Table~\ref{tab:ddpm_ablation}. Together, the t-SNE result and the ablation table support the interpretation that the DDPM branch can provide useful additional local samples under sparse-data conditions, particularly for neighborhood-sensitive classifiers such as KNN.

\subsection{Class-Level Classification and Discrimination Performance}

\begin{figure*}[t]
    \centering
    \includegraphics[width=0.9\linewidth]{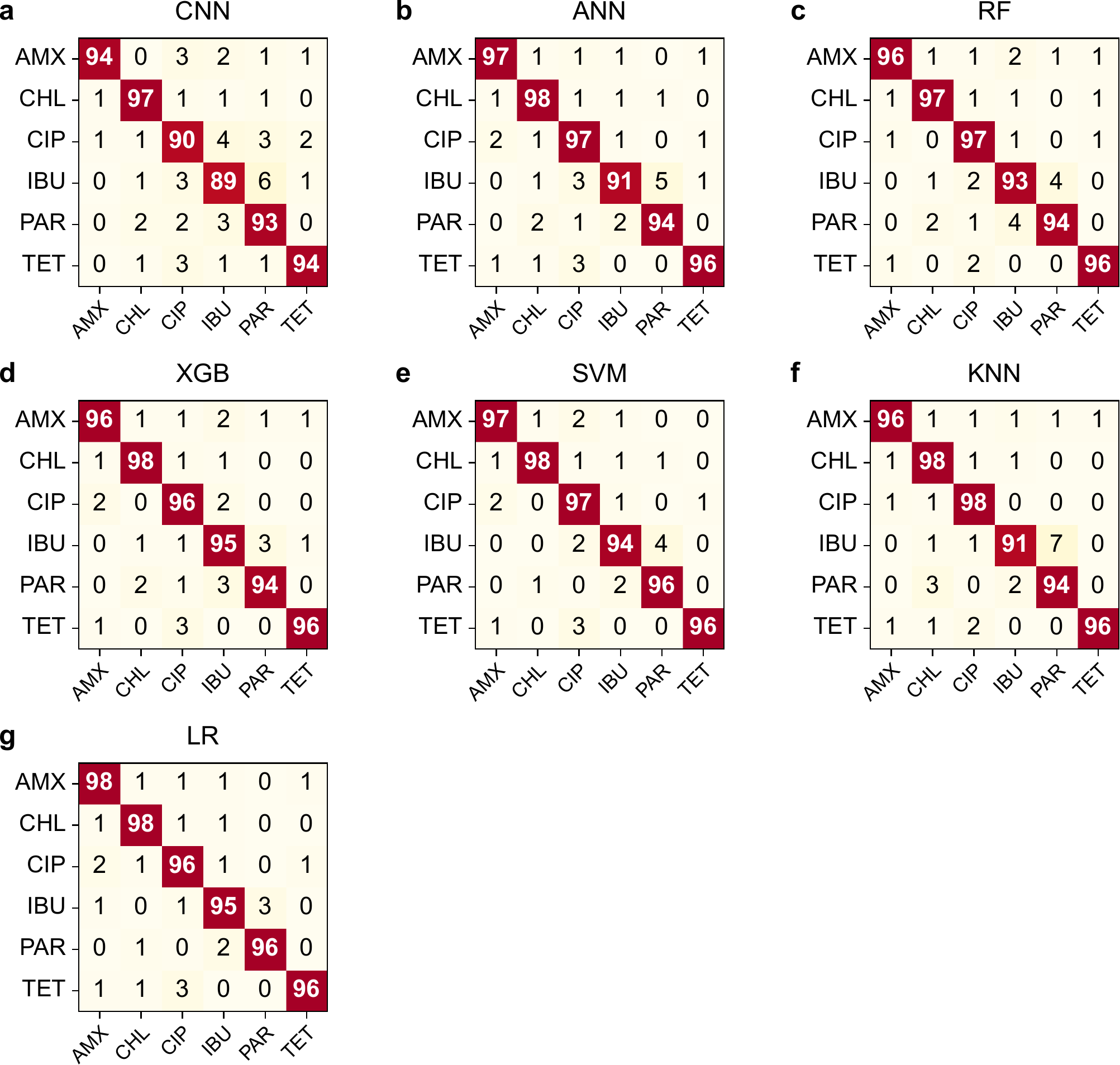}
    \caption{Averaged confusion matrices of the embedding-level CNN baseline and hybrid CNN--ML classifiers under stratified ten-fold cross-validation, where CNN is the EfficientNet-B3 model and ML includes ANN, RF, XGBoost, SVM, KNN, and LR models. Values are reported as class-wise recall percentages. Class labels denote amoxicillin (AMX), chloramphenicol (CHL), ciprofloxacin (CIP), ibuprofen (IBU), paracetamol (PAR), and tetracycline (TET).}
    \label{fig:cm_all_models}
\end{figure*}

In Figures~\ref{fig:cm_all_models}a-\ref{fig:cm_all_models}g, we provide an overview of the averaged confusion matrices for the embedding-level CNN baseline and the hybrid CNN--ML classifiers, where CNN is the EfficientNet-B3 model and ML includes ANN, RF, XGBoost, SVM, KNN, and LR models, respectively. The matrices are normalized on a per-class basis; therefore, each diagonal entry represents the recall of the corresponding pharmaceutical class under stratified ten-fold cross-validation.

The hybrid CNN--ML classifiers generally exhibit stronger diagonal dominance than the CNN baseline. The CNN model still provides reasonable class-level identification, with recall values ranging from 86\% for ibuprofen to 96\% for amoxicillin. However, its off-diagonal errors are more pronounced, particularly for ibuprofen and paracetamol. This pattern is consistent with the lower aggregate performance of the CNN baseline reported in Table~\ref{tab:org_baseline}. In contrast, the hybrid classifiers achieve higher class-wise recall for most compounds, indicating that the EfficientNet-B3 embeddings provide a more discriminative representation when they are coupled with an independent downstream classifier.

\begin{figure*}[t]
    \centering
    \includegraphics[width=0.9\linewidth]{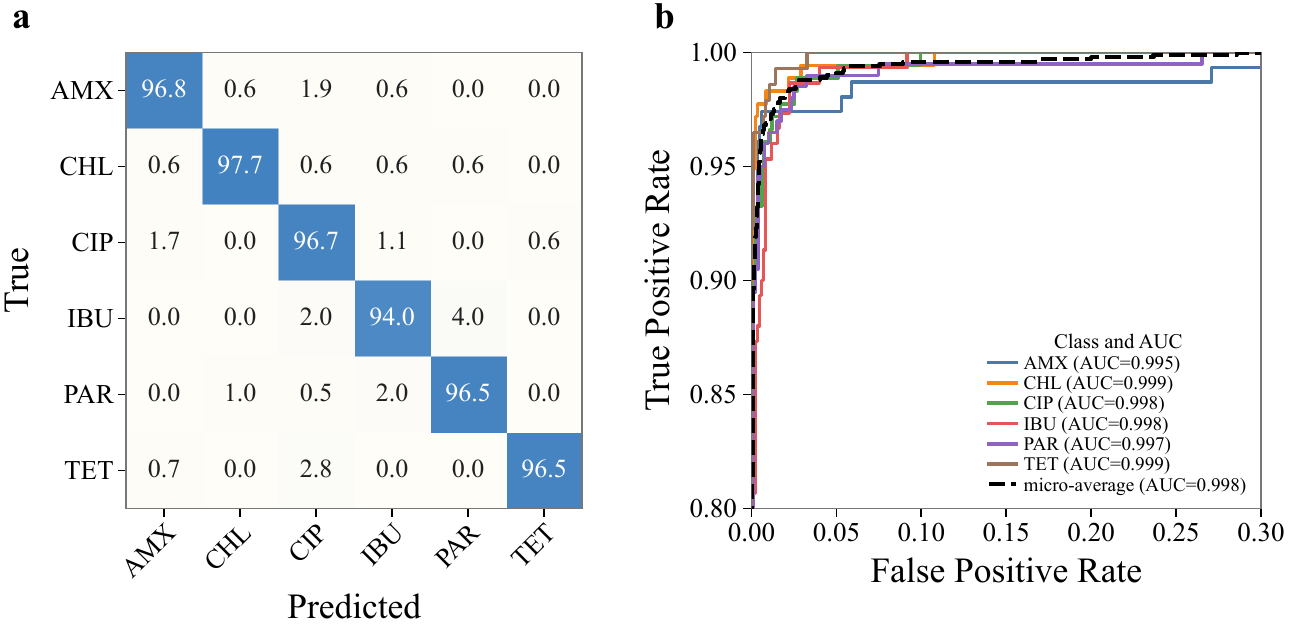}
    \caption{
    Class-level performance of the best-performing EfficientNet-B3--support vector machine (SVM) classifier under stratified ten-fold cross-validation.
    (a) Class-normalized confusion matrix, with values reported as class-wise recall percentages.
    (b) One-vs-rest ROC curves for the six pharmaceutical classes, together with the micro-averaged ROC curve and corresponding AUC values.
}
\label{fig:cm_roc_representative}
\end{figure*}

To examine representative decision mechanisms in greater detail, we calculate the confusion matrices and one-vs-rest ROC--AUC curves of SVM (see Figures~\ref{fig:cm_roc_representative}a-\ref{fig:cm_roc_representative}b), ANN, LR, XGBoost, RF, and KNN (see Figures S3a-S3j). These classifiers represent margin-based, ensemble-based, and distance-based decision principles, respectively. 
Among the six ML classifiers, SVM demonstrates the most balanced class-level behavior, with high recall across all six compounds and a minimum recall of 94.0\% for ibuprofen, as shown in Figure~\ref{fig:cm_roc_representative}a. Its remaining errors are concentrated in a small number of class pairs, while the other classes retain recall values above 95\%. LR (in Figure S3b) and XGBoost (in Figure S3c) show a similarly strong diagonal pattern, although slightly larger off-diagonal entries appear for paracetamol and ciprofloxacin-related predictions. ANN (in Figure S3a) and KNN (in Figure S3e) remain competitive but exhibit more visible confusion between ibuprofen and paracetamol, with a fraction of ibuprofen samples assigned to the paracetamol class. RF (in Figure S3d) performs better for chloramphenicol but worse for amoxicillin, ibuprofen, and paracetamol.

The ROC--AUC results reinforce the patterns observed in the confusion matrices. SVM achieves class-wise AUC values ranging from 0.995 to 0.999, together with a micro-averaged AUC of 0.998, as shown in Figure~\ref{fig:cm_roc_representative}b. ANN, LR, XGBoost, and RF (in Figures S3f, S3g, S3h, and S3i, respectively) provide similarly strong discrimination, with class-wise AUC values from 0.995 to 0.999 and a micro-averaged AUC of 0.997-0.998. These results indicate that the EfficientNet-B3 feature representation clearly separates each target pharmaceutical class from the remaining classes for both margin-based and ensemble-based classifiers. KNN (see Figure S3j) also achieves high class-wise separability, although its ROC performance remains slightly lower than that of other ML models. Its micro-averaged AUC reaches 0.989, while the lowest class-wise AUC values occur for ibuprofen and paracetamol, at 0.981 and 0.989, respectively. This behavior is consistent with the corresponding confusion matrix, in which these two classes show the most prominent residual overlap.

Overall, the confusion-matrix and ROC--AUC analyses provide consistent evidence that the hybrid CNN--ML classifiers maintain strong discrimination across all six pharmaceutical compounds. SVM provides the strongest and most balanced overall baseline, while ANN, LR, XGBoost, and RF remain closely comparable. KNN also performs competitively but is more sensitive to local overlap between spectrally similar samples, particularly ibuprofen and paracetamol. These findings demonstrate that the EfficientNet-B3 feature representation supports reliable class-wise identification and high one-vs-rest separability, rather than producing high aggregate accuracy alone.

\subsection{Application-Level Demonstration Using a Raman Pharmaceutical Analyzer}

To examine the practical applicability of the proposed framework beyond offline model evaluation, we integrate the trained EfficientNet-B3 classifier into a custom Raman Pharmaceutical Analyzer. The application receives a Raman spectrum as input and returns the predicted compound class, the corresponding probability distribution, and a spectral response map for interpretation.

\begin{figure*}[t]
    \centering
    \includegraphics[width=0.9\linewidth]{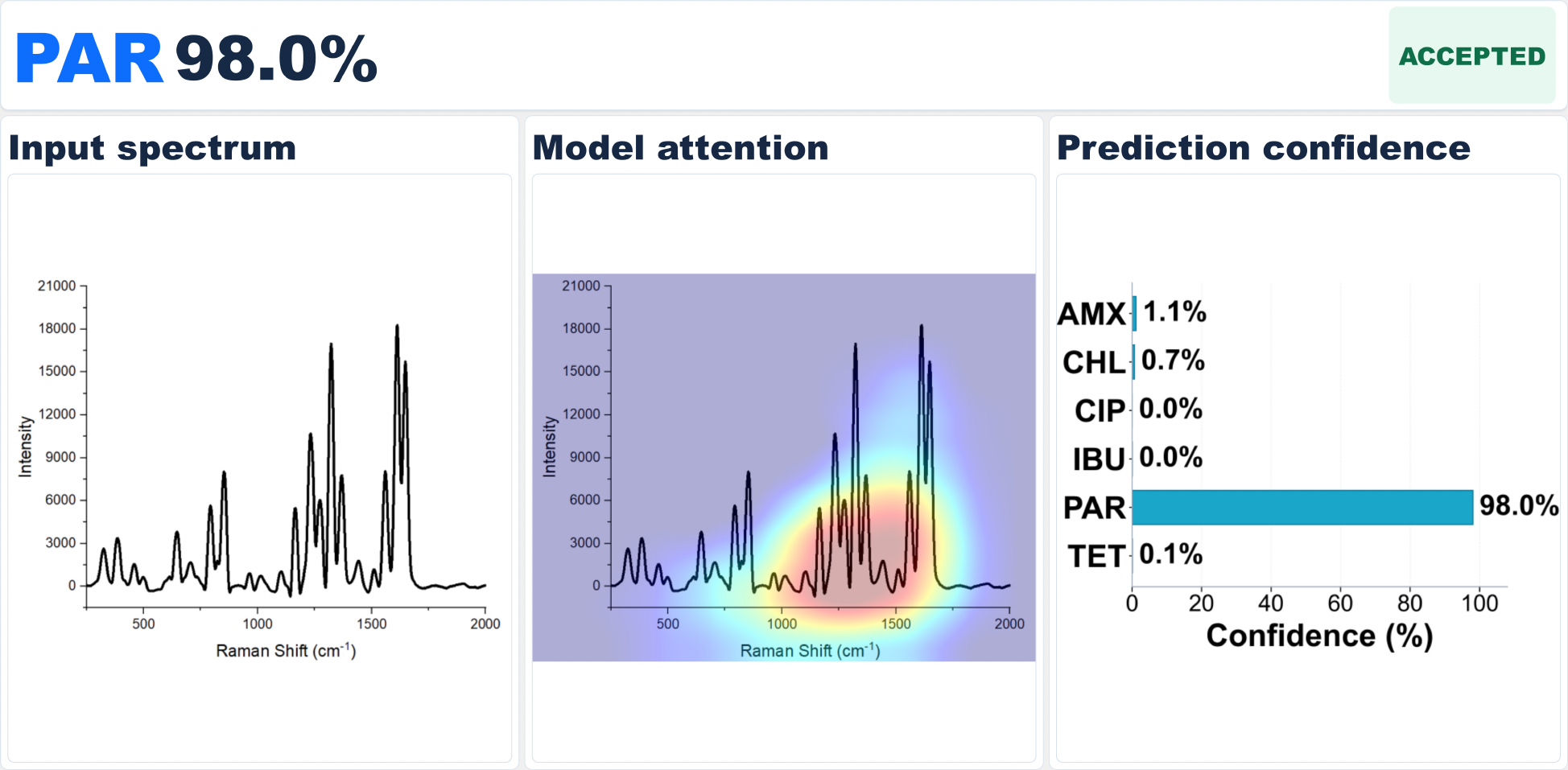}
    \caption{
    Application-level demonstration using the deployed Raman Pharmaceutical Analyzer for an independently measured paracetamol (PAR) sample.
    The interface displays the input Raman spectrum, the model-attention visualization, and the predicted class-probability distribution.
    The sample is correctly identified as PAR with a prediction confidence of 98.0\%. Additional representative results for CIP, IBU, and TET are provided in Figure S4 of the Supporting Information.}
    \label{fig:batch4_demo}
\end{figure*}

In Figure~\ref{fig:batch4_demo}, we show representative inference results for the paracetamol case. Similar results for other compounds, including tetracycline, ciprofloxacin, and ibuprofen, are shown in Figure S4. These examples cover both antibiotic and non-antibiotic pharmaceutical compounds in the dataset. For each test case, the application displays the raw Raman spectrum, the spectral response map highlighting informative regions, and the predicted class probabilities. In the illustrated examples, the target compounds are correctly identified with high confidence, with prediction probabilities exceeding 98\%. It is noted that the Raman spectra used for testing the application are not included in the training dataset and were obtained from independent experiments.

The highlighted spectral regions provide a qualitative indication of the Raman features that contribute to the model decision. These regions are not used as independent chemical evidence, but they help show that the inference output is associated with structured spectral information rather than only a final class label. This behavior is consistent with the class-level confusion matrix and ROC--AUC analyses presented earlier, where the strongest CNN--ML classifiers show high recall and stable class-wise separability.

Overall, the Raman Pharmaceutical Analyzer serves as an application-level demonstration of the proposed framework, showing that the trained EfficientNet-B3 classifier can perform direct inference on experimentally measured Raman spectral images and provide interpretable outputs, including the predicted compound class, confidence distribution, and spectral response map. Although this application does not replace broader external validation on newly collected samples, it demonstrates the feasibility of embedding a trained Raman image classifier into an interactive workflow for rapid pharmaceutical identification.

\section{CONCLUSION}

This study presents HyMLRaman, a hybrid Raman identification framework that combines CNN-derived spectral embeddings with classical machine-learning classifiers for pharmaceutical compound classification. Among the evaluated deep feature extractors, EfficientNet-B3 provides the most effective representation for downstream classification. When its 1536-dimensional embeddings are coupled with classical classifiers, the hybrid CNN--ML pipeline clearly outperforms the corresponding deep-learning baselines. In particular, SVM achieves the strongest overall baseline performance, with 96.31\% accuracy and a macro-F1 score of 96.36\% under stratified ten-fold cross-validation.

The DDPM module is introduced as a feature-level augmentation component for limited-data learning. Rather than generating raw Raman spectra, the model operates in a PCA-reduced EfficientNet-B3 latent space and produces class-conditioned synthetic feature candidates that are retained only after confidence-based and nearest-neighbor filtering. The low-data ablation results show that DDPM augmentation provides selective benefits rather than uniform improvements across all classifiers. The clearest gains are observed for KNN under reduced training fractions, with smaller improvements for SVM in selected settings. These findings indicate that DDPM is most useful as an auxiliary low-data support module, particularly when the original feature distribution is sparsely sampled.

Class-level confusion matrices and ROC--AUC analyses further show that the strongest CNN--ML classifiers maintain stable discrimination across the six pharmaceutical compounds. Five ML models, including SVM, ANN, LR, XGBoost, and RF, exhibit high class-wise AUC values and strong diagonal recall patterns, while KNN remains competitive but is more affected by residual overlap between spectrally similar compounds such as ibuprofen and paracetamol. These results demonstrate that the proposed feature representation supports reliable class-wise identification, not only high aggregate accuracy.

Finally, the Raman Pharmaceutical Analyzer provides an application-level proof-of-concept for deploying the trained model in an interactive Raman analysis workflow. The illustrated inference examples show high-confidence predictions together with spectral response maps that help relate the model output to informative Raman regions. Although further validation on newly collected external samples is still required before practical deployment, the present results show that HyMLRaman provides a promising and interpretable route for rapid Raman-based pharmaceutical screening.

\printcredits

\section*{Declaration of competing interest}
The authors declare that they have no known competing financial interests or personal relationships that could have appeared to influence the work reported in this paper.

\section*{Acknowledgements}
This research is funded by Ho Chi Minh City Department of Science and Technology, Vietnam under contract number 25/2025/HD-QKHCN; through the Project for Developing Mechanisms to Foster the Establishment and Growth of International-Standard Centers of Excellence (CoE Project) as per Decision No. 5721/QD-UBND dated December 11, 2023, of the People Committee of Ho Chi Minh City.


\section*{Supplementary Information}
The Supporting Information contains additional details on the computational efficiency of the HyMLRaman framework and DDPM-based augmentation; fold-wise validation proportions and dataset class distributions; class-level confusion matrices and ROC--AUC analyses of the additional CNN--ML classifiers; and representative outputs of the HyMLRaman prediction interface.

\section*{Data availability}
All scripts, data sets, and source code are available at \url{https://github.com/nguyen-group/HyMLRaman}.

\bibliographystyle{cas-model2-names}

\bibliography{ref}



\end{document}